\documentclass{article}
\usepackage{multirow}
\usepackage{ulem}
\usepackage{subcaption}
\usepackage{microtype}
\usepackage{graphicx}
\usepackage{booktabs} 

\usepackage{hyperref}

\usepackage[accepted]{icml2023}

\usepackage{amsmath}
\usepackage{amssymb}
\usepackage{mathtools}
\usepackage{amsthm}

\usepackage[capitalize,noabbrev]{cleveref}

\theoremstyle{plain}

\theoremstyle{definition}

\theoremstyle{remark}

\usepackage[textsize=tiny]{todonotes}

\icmltitlerunning{}

\begin{document}

\twocolumn[
\icmltitle{Advanced Knowledge Transfer: Refined Feature Distillation for Zero-Shot Quantization in Edge Computing}



\icmlsetsymbol{equal}{*}

\begin{icmlauthorlist}
\icmlauthor{Inpyo Hong}{1}
\icmlauthor{Youngwan Jo}{1}
\icmlauthor{Hyojeong Lee}{2}
\icmlauthor{Sunghyun Ahn}{1}
\icmlauthor{Sanghyun Park}{1}
\end{icmlauthorlist}

\icmlaffiliation{1}{Department of Computer Science, Yonsei University, Seoul, Republic of Korea}
\icmlaffiliation{2}{Department of Artificial Intelligence, Yonsei University, Seoul, Republic of Korea}

\icmlcorrespondingauthor{Sanghyun Park}{sanghyun@yonsei.ac.kr}

\icmlkeywords{Machine Learning, ICML}

\vskip 0.3in
]



\printAffiliationsAndNotice{}  

\begin{abstract}
We introduce AKT (Advanced Knowledge Transfer), a novel method to enhance the training ability of low-bit quantized (Q) models in the field of zero-shot quantization (ZSQ). Existing research in ZSQ has focused on generating high-quality data from full-precision (FP) models. However, these approaches struggle with reduced learning ability in low-bit quantization due to its limited information capacity. To overcome this limitation, we propose effective training strategy compared to data generation. Particularly, we analyzed that refining feature maps in the feature distillation process is an effective way to transfer knowledge to the Q model. Based on this analysis, AKT efficiently transfer core information from the FP model to the Q model. AKT is the first approach to utilize both spatial and channel attention information in feature distillation in ZSQ. Our method addresses the fundamental gradient exploding problem in low-bit Q models. Experiments on CIFAR-10 and CIFAR-100 datasets demonstrated the effectiveness of the AKT. Our method led to significant performance enhancement in existing generative models. Notably, AKT achieved significant accuracy improvements in low-bit Q models, achieving state-of-the-art in the 3,5bit scenarios on CIFAR-10. The code is available at \textit{https://github.com/Inpyo-Hong/AKT-Advanced-knowledge-Transfer}.\looseness=-1
\end{abstract}

\section{Introduction}
\label{sec:intro}

In recent years, deep neural networks (DNNs) have advanced across a wide range of fields, including natural language processing, computer vision, and multi-modal learning\cite{32, 33, 34, 35}. Despite this progress, the need for significant computational resources in DNNs remains a key challenge to overcome\cite{23, 59}. In particular, it makes hard to utilizing these models efficiently on edge devices, such as mobile devices and the Internet of Things (IoT)\cite{44, 45, 46, 47}. To address this challenge, the DNN field has been actively exploring techniques to compress the models\cite{48, 49} or their utilized data\cite{24, 50}. In more detail, recent DNNs compression research can be categorized into model compression\cite{21} and data compression\cite{24}. In the area of model compression, significant research has been conducted in knowledge distillation\cite{1}, pruning\cite{2}, and quantization\cite{3}. Among these, quantization stands out as a powerful method to maximize DNNs efficiency, reducing both latency and computing resource consumption\cite{4, 5, 6, 60, 61}.


\sloppy

Quantization techniques are classified based on the timing of quantization into quantization-aware training (QAT)\cite{7} and post-training quantization (PTQ)\cite{8}. While QAT applies quantization during the training process to maintain high performance, it requires full access to the training data. This requirement can limit its practicality\cite{27, 56, 57, 58}. On the other hand, PTQ is simpler to implement but carries the risk of performance degradation\cite{27, 51, 52, 53}. Additionally, PTQ presents challenges in environments with sensitive data, like healthcare and finance, where further training is restricted\cite{54, 55}.


To address these limitations, zero-shot quantization (ZSQ), or data-free quantization, has emerged as a key technique that performs quantization without access to training data\cite{9, 10, 62, 63}. ZSQ ensures data independence and is highly applicable across various model architectures, making it an attractive approach. As indicated in Table \ref{tab1}, current ZSQ research is primarily directed toward data generation\cite{13, 14, 15, 16, 17, 18, 19, 20} rather than training methods\cite{11, 12}. These generation methods approaches primarily aim to generate high-quality data by improving the training strategies of generators. Data generation studies have achieved nearly full-precision model performance in 5-bit quantization and remarkable performance improvements in 4-bit settings\cite{18, 19, 20}. However, in contrast to 4-bit and 5-bit, low-bit environments such as 3-bit still suffer from performance limitations even with sufficient training\cite{26}. This indicates that focusing solely on data generation may not be sufficient, and a more fundamental learning-based approach is required.


\begin{table}[]
\renewcommand{\arraystretch}{1.3}
\resizebox{0.45\textwidth}{!}{%
\begin{tabular}{cc}
\hline
Data Generation Methods                                                                                                                                   & Training Methods                                                                          \\ \hline
\begin{tabular}[c]{@{}c@{}} 
{\small DSG \textit{(CVPR 21) \cite{13}}}\\ 
{\small Qimera \textit{(AAAI 21) \cite{14}}}\\ 
{\small IntraQ \textit{(CVPR 22) \cite{15}}}\\ 
{\small HAST \textit{(CVPR 23) \cite{16}}}\\ 
{\small Ada SG \textit{(AAAI 23) \cite{17}}}\\ 
{\small Ada DFQ \textit{(CVPR 23) \cite{18}}}\\
{\small TexQ \textit{(NeurIPS 24) \cite{19}}}\\ 
{\small RIS \textit{(AAAI 24) \cite{20}}} 
\end{tabular} 
& 
\begin{tabular}[c]{@{}c@{}} 
{\small SQuant \textit{(ICLR 22) \cite{11}}}\\ 

{\small AIT \textit{(CVPR 22) \cite{12}}} 
\end{tabular} 
\\ \hline
\end{tabular}%
}
\caption{Categorization of Zero-shot Quantization Algorithms}
\label{tab1}
\end{table}

Therefore, we thoroughly examined the factors contributing to performance drops in low-bit quantization. Firstly, the limited capacity of low-bit quantized models reduces the efficiency of conventional distillation methods. This inefficiency may result in gradient explosion issues. Secondly, the current knowledge distillation strategies used in ZSQ are not optimal for training low-bit quantized models with their limited capacity.\\


Based on this analysis, we propose AKT (Advanced Knowledge Transfer), which aims to enhance the training efficiency of low-bit quantized models. Unlike conventional feature distillation, AKT effectively captures both channel and spatial attention information from the feature maps of full-precision models. Through this, it ensures that key features are effectively incorporated into the training process. By applying our method to various generative methods, we have achieved significant performance enhancement, especially in low-bit environments like 3-bit quantization. The key contributions of our work are summarized below:


\begin{itemize}
     \item We identified the critical role of both channel and spatial information in training zero-shot quantized models. Our analysis using second-order curvature shows that preserving these components is crucial for maintaining performance during quantization.
    \item We observed that integrating channel and spatial information significantly improves the performance of the quantized model. To the best of our knowledge, this is the first work to systematically analyze and highlight the importance of these components in zero-shot quantization.
    \item Based on these findings, we propose the AKT (Advanced Knowledge Transfer) method, which proves highly effective for training low-bit quantized models and achieves state-of-the-art performance, particularly in the context of 3-bit and 5-bit zero-shot quantization.
    \item Our proposed AKT method demonstrates broad applicability, improving diverse data generation methods and advancing the development of learning strategies in zero-shot quantization.

\end{itemize}

\section{Related Work}
\label{sec:relatedwork}

\subsection{Quantization}

Quantization \cite{25} converts continuous real values into a set of discrete integer values. It is divided into uniform and non-uniform quantization based on interval spacing and into symmetric and asymmetric quantization based on the symmetry of the quantization range \cite{21}. In this study, we uses asymmetric quantization, which is better suited for image processing tasks where data tends to cluster within a specific range. The quantization process is detailed in Equations \eqref{eq1},\eqref{eq2}, and \eqref{eq3} \cite{27}.


\begin{equation}
     s = \frac{x_{max}-x_{min}}{2^{k}-1}, z = \left [ \frac{-x_{min}}{s} \right ] 
\label{eq1}
\end{equation}

\begin{equation}
    q = \left [ \frac{x}{s} + z \right ] 
\label{eq2}
\end{equation}

\begin{equation}
    \hat{x} = s \cdot (q - z)
\label{eq3}
\end{equation}

\begin{figure*}[t]
    \centering
    \includegraphics[width=0.98\textwidth, height=3.6in]{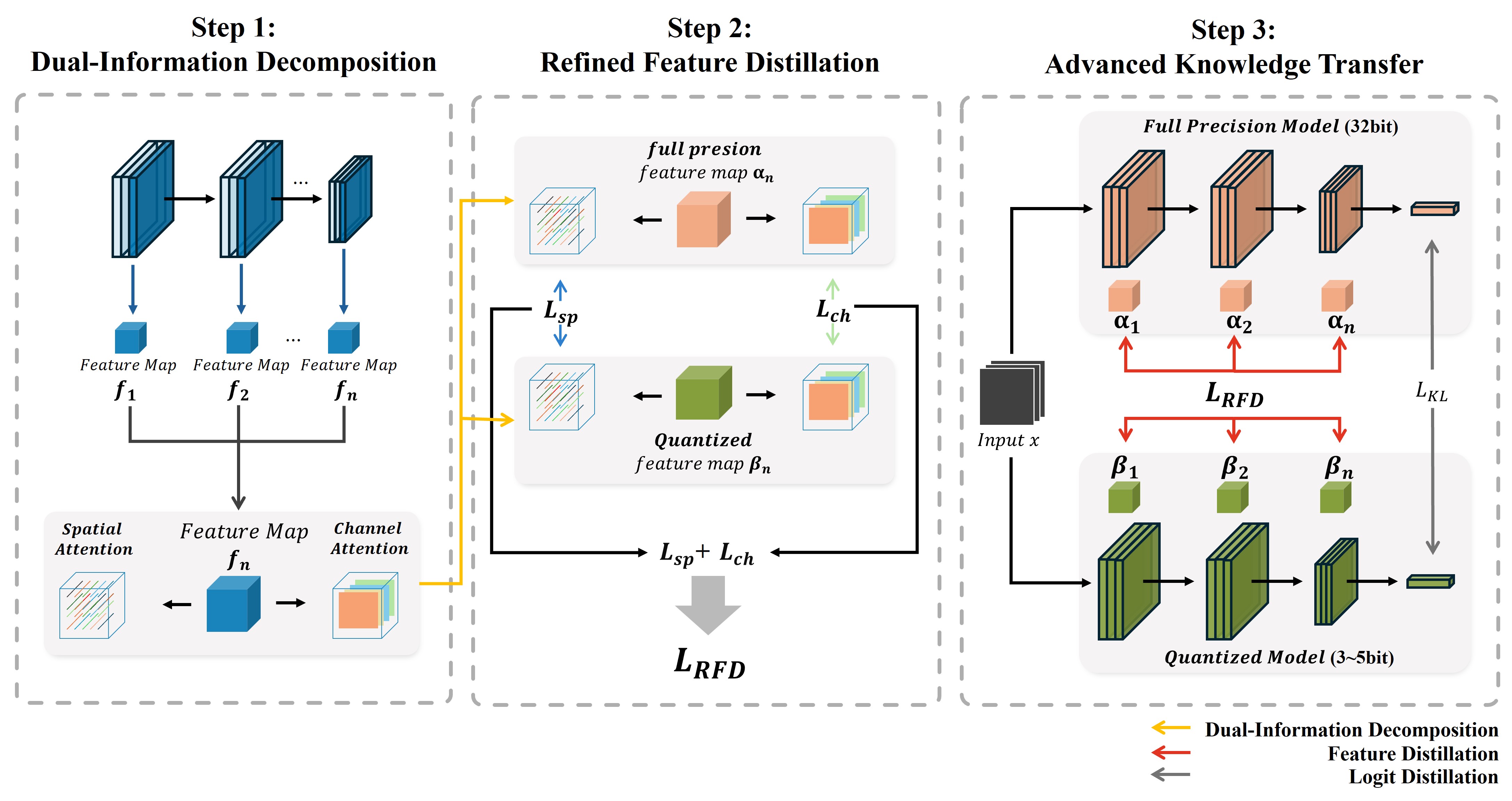}
    \caption{An overview of AKT(Advanced Knowledge Transfer) method. Step 1 illustrates the process of decomposing each feature map into spatial and channel information. Step 2 demonstrates the computation of the refined `RFD loss' through the integration of spatial and channel losses. Step 3 presents the process of transferring the enhanced feature knowledge into the quantized model using `RFD loss' in a zero-shot quantization setting.}
    \label{fig1}
\end{figure*}

To perform quantization, Equation \eqref{eq1} is first used to calculate the scale $s$ and zero-point $z$ within the specified range [$x_{min}, x_{max}$]. $x$ denotes input, and the parameter $k$ represents the bit-width utilized for quantization. Following this, Equation \eqref{eq2} is employed to quantize the floating-point value $x$ into an integer value $q$. Finally, dequantization interprets the final output by reconstructing $\hat{x}$ from the quantized integer $q$ in Equation \eqref{eq3}.


\subsection{Zero-shot Quantization}

Conventional quantization methods depend on training data to maintain model performance. However, ZSQ is designed to effectively preserve quantization performance without any training data\cite{9}. Xu et al.\cite{22} first introduced GDFQ, which trains quantized models through data generation in the field of ZSQ. Following this work, research has evolved into two main branches: data generation and training methodologies, as summarized in Table \ref{tab1}.


\subsubsection{Data Generation Methodology}

Data generation is pivotal in ZSQ, particularly when training data is not explicitly available. GDFQ\cite{22} introduced a method for generating data based on gaussian distribution. This approach has been widely adopted as a baseline in many studies. DSQ generates data by leveraging model structure within the feature space\cite{13}. Qimera further refined DSQ's approach, suggesting a strategy to cover a broader data distribution by exploring multiple paths\cite{14}. While recent advancements have focused on developing more sophisticated data generation techniques to improve quantization performance\cite{15, 16, 18, 19, 20}, they have encountered performance limitations in low-bit quantization. This highlights the necessity of exploring alternative approaches to enhance performance in such environments.


\subsubsection{Training Methodology}

Compared to data generation, research on training methods in ZSQ has been relatively underdeveloped, but their importance is becoming more prominent. Although data generation is crucial for improving quantization performance, the effectiveness of a training approach can be maximized when applicable to various generation methods. This indicates that a training approach is adaptable and not restricted to any particular data generation method. This allows for better quantization performance when combined with different techniques. SQuant\cite{11} utilized diagonal Hessian approximation to minimize the performance degradation in the quantization process and emphasized the need for data-independent training methods. AIT\cite{12} examined the negative impact of combining cross-entropy loss with KL loss, underlining the importance of selecting an appropriate loss function for ZSQ. Nevertheless, existing research has not addressed the challenges of effectively training low-bit quantized models. To tackle this issue, we propose a novel approach.


\section{Preliminaries}
\label{sec:preliminaries}
\sloppy
\subsection{Knowledge Distillation}
Knowledge distillation (KD) \cite{1} is currently the most widely adopted training approach in the ZSQ field. KD is actively explored from various perspectives, such as learning principles and security considerations \cite{28, 29, 30, 31, 64}, it is still widely utilized in ZSQ. KD transfers knowledge from a teacher model to a smaller student model. This process is categorized as either logit or feature distillation. The equations for these distillations are presented in Equations \eqref{eq4} and \eqref{eq5}, respectively.


\begin{equation}
    \mathcal{L}_{\text{logit distillation}} = \tau^2 \cdot \text{KL}\left( \sigma\left(\frac{T(x)}{\tau}\right) \| \sigma\left(\frac{S(x)}{\tau}\right) \right)
    \label{eq4}
\end{equation}

\begin{equation}
    \mathcal{L}_{\text{feature distillation}} = d(f_T(x), f_S(x))
\label{eq5}
\end{equation}

In Equation \eqref{eq4}, $T(x)$ and $S(x)$ correspond to the logits of the teacher network and student network, respectively. $\tau$ represents the temperature parameter used in distillation. Additionally, $\sigma$ denotes the softmax function and $KL$ stands for the kullback-leibler divergence. Thus, the purpose of logit distillation is to reduce the discrepancy between the outputs of the teacher and student networks. Temperature parameter $\tau$ smooth the logits to enhance the learning performance of the student network.
In the case of feature distillation represented by Equation \eqref{eq5}, $f_T(x)$ and $f_S(x)$ refer to the feature maps of the teacher and student networks. The term $d$ stands for the distance between the losses of individual features. Therefore, it focuses on transferring knowledge from the internal feature information of the network, rather than the logits. 


In the current field of ZSQ, the primary focus has been on logit distillation, which relies on the model's final output (logit). However, it may lead to information loss in the model\cite{12}. To address this issue, feature distillation can be applied to effectively reflect the feature information of the model\cite{36}. By utilizing features extracted from the intermediate layers, feature distillation helps maintain enhanced quantization performance without the need for training data. This helps preserve the crucial information of the full-precision model in ZSQ.


\section{AKT Method}
\label{sec:akt method}

We propose a novel AKT method to improve the performance of low-bit quantized models in a ZSQ. As shown in Figure \ref{fig1}, the AKT consists of 3 stages, which will be detailed in this section.

\begin{figure}[t]
    \centering
    \includegraphics[width=0.38\textwidth, height=3.5in]{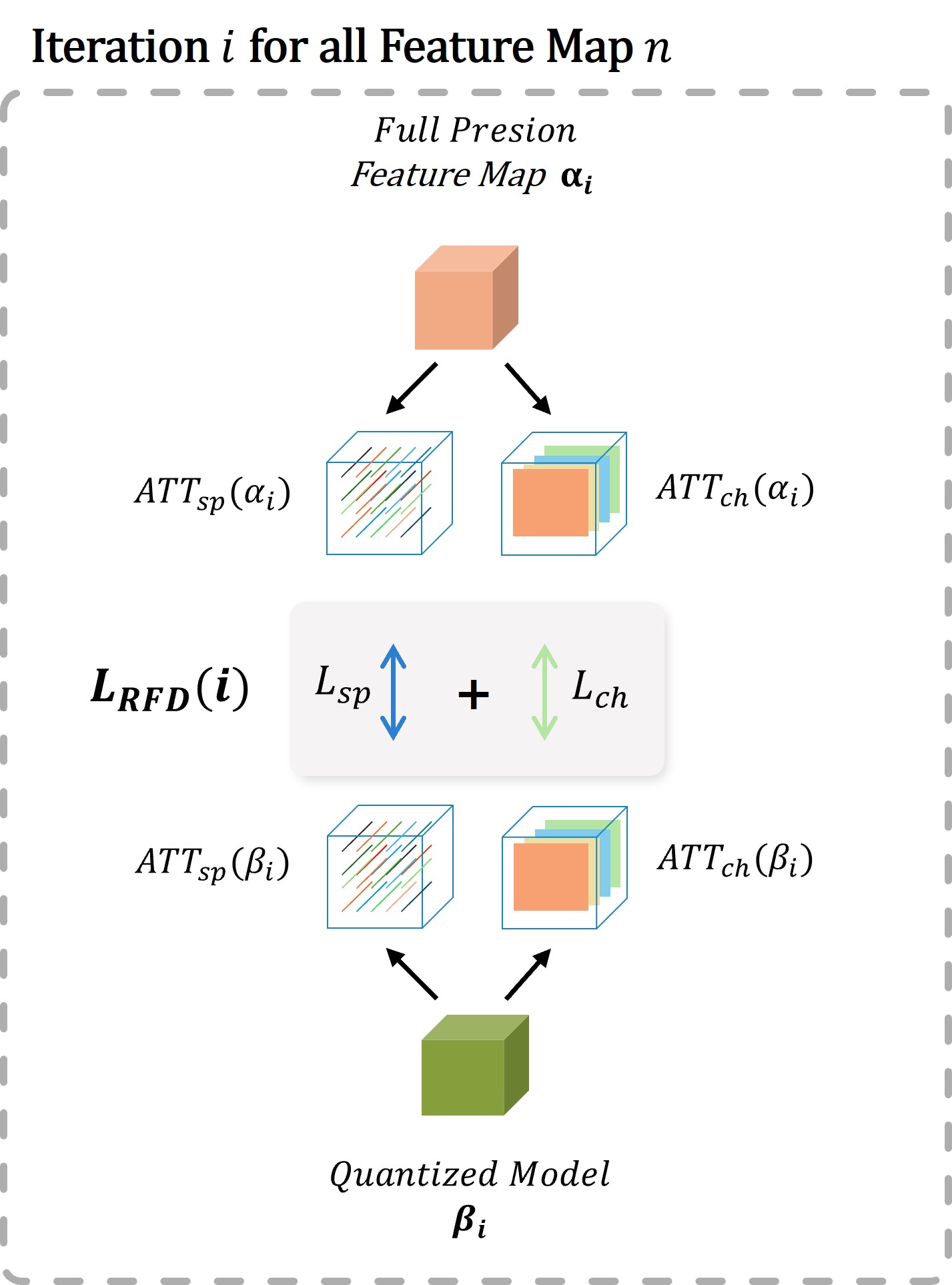}
    \caption{An illustration of Refined Feature Distillation. The distillation process is independently applied to each of the $n$ layers, and the resulting losses are averaged to compute the final $L_{RFD}$loss.}
    \label{fig2}
\end{figure}

\subsection{Dual-Information Decomposition}

The process of dual-information decomposition is illustrated in the first step of Figure \ref{fig1}. In this stage, feature maps (i.e., activation maps) are extracted from various layers of the model, similar to the basic feature distillation process. Subsequently, we apply both spatial and channel attention to each feature map to extract key information. Equations \eqref{eq6} and \eqref{eq7} represent the spatial and channel attention, respectively.


\begin{equation}
ATT_{sp}(x) = \frac{\frac{1}{C} \sum_{c=1}^C x_c^2}{\left\|\frac{1}{C} \sum_{c=1}^C x_c^2\right\|}, \text{ where }  x_c \in \mathbb{R}^{H \times W}
\label{eq6}
\end{equation}

\begin{equation}
ATT_{ch}(x) = \sigma \left( \frac{1}{H \cdot W} \sum_{h=1}^H \sum_{w=1}^W x_{h,w}^2 \right), \text{ where } x_{h,w} \in \mathbb{R}^{C}
\label{eq7}
\end{equation}

\begin{table*}[t]
\centering
\renewcommand{\arraystretch}{1.7}
\resizebox{0.95\textwidth}{!}{%

\begin{tabular}{c|c|c|cc|cc|cc}
\toprule
\hline
\textbf{Dataset} & \textbf{\begin{tabular}[c]{@{}c@{}}Model\\ (Full-precision)\end{tabular}} & \textbf{Bits} & \multicolumn{1}{c|}{\textbf{\begin{tabular}[c]{@{}c@{}}GDFQ\\ \textit{(ECCV 20)}\end{tabular}}} & \textbf{\begin{tabular}[c]{@{}c@{}}Ours\\ \textit{(GDFQ+AKT)}\end{tabular}} & \multicolumn{1}{c|}{\textbf{\begin{tabular}[c]{@{}c@{}}HAST$^\dagger$\\ \textit{(CVPR 23)}\end{tabular}}} & \textbf{\begin{tabular}[c]{@{}c@{}}Ours\\ \textit{(HAST+AKT)}\end{tabular}} & \multicolumn{1}{c|}{\textbf{\begin{tabular}[c]{@{}c@{}}AdaDFQ\\ \textit{(CVPR 23)}\end{tabular}}} & \textbf{\begin{tabular}[c]{@{}c@{}}Ours\\ \textit{(AdaDFQ+AKT)}\end{tabular}} \\ \hline
\textbf{} & \textbf{} & 3w3a & 72.23$^\dagger$ & 72.80(+0.57) & 84.11 & 84.84(+0.73) & 84.89 & \textbf{86.76(+1.87)} \\
 & & 4w4a & 90.25 & 90.33(+0.08) & 92.00 & 92.25(+0.25) & 92.31 & \textbf{92.64(+0.33)} \\ 
\multirow{-3}{*}{CIFAR-10} & \multirow{-3}{*}{\begin{tabular}[c]{@{}c@{}}ResNet-20\\ 93.89\end{tabular}} & 5w5a & 93.38 & 93.36(-0.02) & 93.48 & 93.57(+0.09) & 93.81 & \textbf{93.83(+0.02)} \\ \hline
\textbf{} & \textbf{} & 3w3a & 47.59$^\dagger$ & 49.06(+1.47) & 45.21 & 46.80(+1.59) & 52.74 & \textbf{54.68(+1.94)} \\
 & & 4w4a & 63.39 & 63.37(-0.02) & 64.40 & 65.26(+0.86) & 66.81 & \textbf{66.94(+0.13)} \\ 
\multirow{-3}{*}{CIFAR-100} & \multirow{-3}{*}{\begin{tabular}[c]{@{}c@{}}ResNet-20\\ 70.33\end{tabular}} & 5w5a & 66.12 & 67.46(+1.34) & 68.82 & 68.91(+0.09) & \textbf{69.93} & \ 69.75(-0.18) \\ \hline \bottomrule

\end{tabular}%
}
\caption{Top-1 accuracy comparison of AKT with zero-shot quantization methods. In $n$w$m$a, $n$ and $m$ represent the quantization bits for weights and activations, respectively. The best performance is in bold, and $\dagger$ denotes our re-implementation.}
\label{tab2}
\end{table*}

$x$, $C$, $H$, and $W$ represent the feature map, channel, height, and width, respectively in Equations \eqref{eq6}, \eqref{eq7}. $x_c^2$ in Equation \eqref{eq6} is the square of the feature map values for each channel $c$. Spatial attention aggregates all channel values and then divides by the total number of channels $C$, thereby preserving the width and height information while reducing the channel dimension by averaging its values. Subsequently, $L_2$ normalization emphasizes spatial pixel information, as described in Equation \eqref{eq6}. Similarly, the channel attention in Equation \eqref{eq7} preserves channel information by averaging the spatial dimensions $h$ and $w$. This process allows for calculating the relative importance of each channel, which is then normalized using softmax, represented by $\sigma$. 

By utilizing these two equations, the feature map is refined from diverse perspectives. This enables the effective transfer of high-quality information to quantized models with limited information capacity.



\subsection{Refined Feature Distillation}

Based on dual-information decomposition (step 1), we propose refined feature distillation loss ($L_{RFD}$), which integrates refined feature information from different perspectives.


\begin{equation}
\begin{split}
L_{RFD} = \lambda \times \frac{1}{N} \sum_{i=1}^{N} \Big( KL \left( ATT_{sp} (\alpha_i) \parallel ATT_{sp} (\beta_i) \right) \\
+ KL \left( ATT_{ch} (\alpha_i) \parallel ATT_{ch} (\beta_i) \right) \Big)
\end{split}
\label{eq8}
\end{equation}

The $KL$ function denotes the kullback-leibler divergence loss function. $\alpha_{i}$ denotes the feature map from the full-precision model, and $\beta_{i}$ represents the corresponding feature map from the quantized model. The first $KL$ term, $KL( ATT_{sp} (\alpha_i) \parallel ATT_{sp} (\beta_i))$, represents the spatial attention loss $L_{sp}$. This loss calculates the spatial feature differences between the teacher model (full-precision model) and the student model (quantized model).  Similarly, the second $KL$ term, $KL( ATT_{ch} (\alpha_i) \parallel ATT_{ch} (\beta_i))$, corresponds to the channel attention loss $L_{ch}$, which measures the differences in channel-focused features. Adding $L_{sp}$ and $L_{ch}$ gives the $L_{RFD}(i)$ for the $i$-th feature map, as illustrated in Figure \ref{fig2}. The hyperparameter $\lambda$ amplifies the effect of feature distillation, and $N$ denotes the number of feature maps used for distillation. The final $L_{RFD}$ is computed as the mean of the $N$ feature maps, as expressed in Equation \eqref{eq8}.

Since feature distillation involves multiple feature map operations, it has certain computational resource limitations. Therefore, set the number of feature maps for distillation, $N$, according to the number of stages in the model. We then utilize the feature maps obtained from the final layer of each module for distillation.



\subsection{Advanced Knowledge Transfer}
Ultimately, we propose the AKT Method, which builds upon the commonly used logit distillation in ZSQ by adding our newly designed $L_{RFD}$. As depicted in Step 3 of Figure \ref{fig1}, the final equation applied to train the quantized model is described in Equation \eqref{eq9}.


\begin{equation}
L_{AKT} = \alpha L_{RFD} + (1-\alpha)L_{KL}
\label{eq9}
\end{equation}

Here, $\alpha$ determines the ratio between the $L_{RFD}$ and the $L_{KL}$. By default, we set it to 0.5. Given that ZSQ involves a diverse range of generated data, the training process often requires adjustments based on the data's characteristics. Thus, users can adjust $\alpha$ dynamically to better suit the generated data. Although we apply the standard kullback-leibler divergence loss ($L_{KL}$) for logit distillation, this component can also be modified depending on the properties of the generated data.


Through $L_{AKT}$, we can transfer both the feature information and output information of the full-precision model to the quantized model in a ZSQ environment. This approach is a key factor in enhancing the learning capabilities of the quantized model.


\section{Experiments}
\label{sec:experiments}

\begin{table}[t]
\centering
\renewcommand{\arraystretch}{1.3}
\resizebox{\columnwidth}{!}{%
\begin{tabular}{c|c|ccc}
\toprule
\hline
\multirow{2}{*}{\begin{tabular}[c]{@{}c@{}}\textbf{Dataset}\\ \textbf{(FP32 Acc)}\end{tabular}} & \multirow{2}{*}{\textbf{Method}} & \multicolumn{3}{c}{\textbf{Top-1 Accuracy(\%)}}           \\ \cline{3-5} 
                                                                              &                         & \textbf{3w3a}           & \textbf{4w4a}           & \textbf{5w5a}           \\ \hline
\multirow{7}{*}{\begin{tabular}[c]{@{}c@{}}CIFAR-10\\ (93.89)\end{tabular}}   & GDFQ \textit{(ECCV 20)}          & 72.23$^\dagger$              & 90.25          & 93.38          \\
                                                                              & AIT \textit{(CVPR 22)}           & -              & 91.23          & 93.43          \\
                                                                              & HAST \textit{(CVPR 23)}$^\dagger$          & 84.11          & 92.00          & 93.48          \\
                                                                              & AdaDFQ \textit{(CVPR 23)}        & 84.89          & 92.31          & {\uline{93.81}}    \\
                                                                              & TexQ \textit{(NeurlPS 24)}       & {\uline{86.47}}    & \textbf{92.68} & -              \\
                                                                              & RIS \textit{(AAAI 24)}           & -              & 92.59          & 93.59          \\
                                                                              & AdaDFQ+AKT \textit{(Ours)}              & \textbf{86.76} & {\uline{92.64}}    & \textbf{93.83} \\ \hline
\multirow{7}{*}{\begin{tabular}[c]{@{}c@{}}CIFAR-100\\ (70.33)\end{tabular}}  & GDFQ \textit{(ECCV 20)}          & 47.59$^\dagger$              & 63.39          & 66.12          \\
                                                                              & AIT \textit{(CVPR 22)}           & -              & 65.80          & 69.26          \\
                                                                              & HAST \textit{(CVPR 23)}$^\dagger$          & 45.21          & 64.40          & 68.82              \\
                                                                              & AdaDFQ \textit{(CVPR 23)}        & 52.74          & 66.81          & \textbf{69.93}          \\
                                                                              & TexQ \textit{(NeurlPS 24)}       & \textbf{55.87}          & \textbf{67.18} & -              \\
                                                                              & RIS \textit{(AAAI 24)}           & -              & 65.99          & 69.55          \\
                                                                              & AdaDFQ+AKT \textit{(Ours)}              & \uline{54.68}          & {\uline{66.94}}    & \uline{69.75}          \\ \hline
                                                                              \bottomrule
\end{tabular}%
}
\caption{Comparison of zero-shot quantization methods on CIFAR-10 and CIFAR-100. 
The best performance is in bold, and the second-best is underlined. $\dagger$ denotes our re-implementation. AKT was applied to the latest method for which official code is available.}
\label{tab3}
\end{table}

\begin{table}[b]
\centering
\renewcommand{\arraystretch}{1.2}
\resizebox{\columnwidth}{!}{%
\begin{tabular}{ccc|ccc}
\toprule
\hline
\multicolumn{3}{c|}{\textbf{Feature Distillation Method}}                                                                                       & \multicolumn{3}{c}{\textbf{Top-1 Accuracy(\%)}} \\ \hline
\begin{tabular}[c]{@{}c@{}}\textbf{Channel}\\ \textbf{Attention}\end{tabular} & \begin{tabular}[c]{@{}c@{}}\textbf{Spatial}\\ \textbf{Attention}\end{tabular} & \textbf{RFD} & \textbf{3w3a}      & \textbf{4w4a}      & \textbf{5w5a}     \\ \hline
\checkmark                                                           &                                                             &     & 86.09     & 92.36     & 93.62    \\
                                                            & \checkmark                                                           &     & 86.10     & 92.49     & 93.75    \\
                                                            &                                                             & \checkmark   & \textbf{86.76}     & \textbf{92.64}     & \textbf{93.83}    \\ \hline
\bottomrule
\end{tabular}%
}
\caption{Ablations on various types of attention information for feature distillation. "RFD" denotes our proposed refined feature distillation. The experiments were conducted using the AdaDFQ method and the CIFAR-10 dataset.}
\label{tab4}
\end{table}

\subsection{Experimental Environments}

We evaluated our AKT method on CIFAR-10 and CIFAR-100, which are the most commonly used datasets in network quantization, particularly in ZSQ\cite{37}. For model selection, we also followed standard practice in ZSQ by using the widely used pre-trained ResNet-20\cite{38}. The pre-trained ResNet-20 model was obtained from the PyTorch library\cite{39}.

\sloppy

For a comparative evaluation of our proposed method, we utilized popular ZSQ methods such as GDFQ\cite{22}, HAST\cite{16}, AdaDFQ\cite{18}. Additionally, to verify the superior performance of the AKT method, we included recent state-of-the-art approaches for a more comprehensive quantitative comparison (e.g., AIT\cite{12}, TexQ\cite{19}, RIS\cite{20}). All experiments were conducted on an NVIDIA RTX3090.
The generator’s default settings utilized a learning rate of 1e-3 and the Adam optimizer. For training models with the AKT method, Nesterov SGD was applied with a momentum of 0.9 and a weight decay of 1e-4. The learning rate was adjusted to decrease by a factor of 0.1 every 100 epochs. For logit distillation and feature distillation, the temperatures were set to 20 and 8, respectively. Models for CIFAR-10 and CIFAR-100 were trained with batch sizes of 200 and 16, respectively, over a total of 200 epochs. Since GDFQ did not include experiments in a 3-bit setting, and HAST utilized a full-precision model with different performance, we conducted our own re-implementation. 




\begin{figure}[t]
    \centering

    \begin{subfigure}[t]{\columnwidth}
        \centering
        \includegraphics[width=\textwidth]{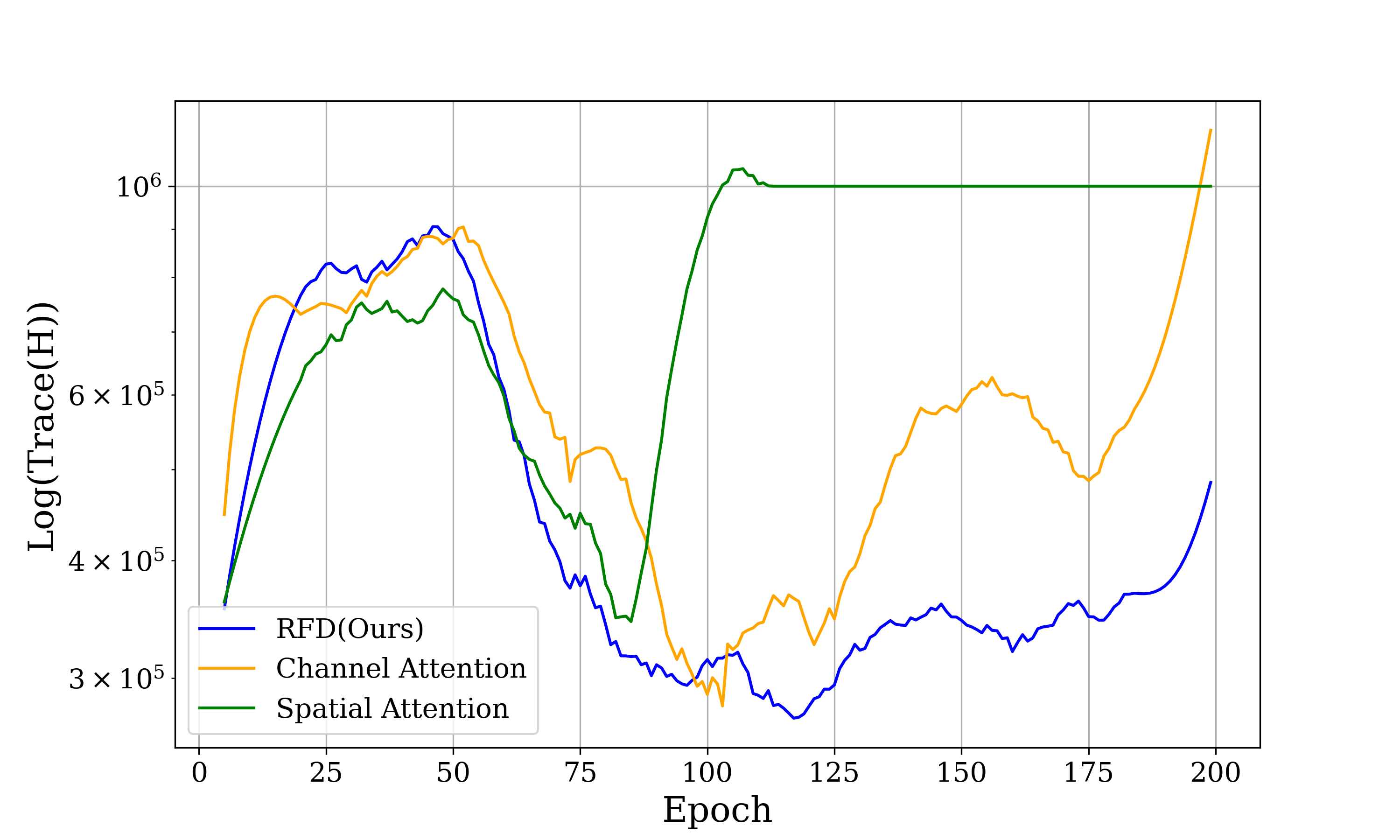}
        \caption{3-bit Quantization}
        \label{fig3a}
    \end{subfigure}

    \begin{subfigure}[t]{\columnwidth}
        \centering
        \includegraphics[width=\textwidth]{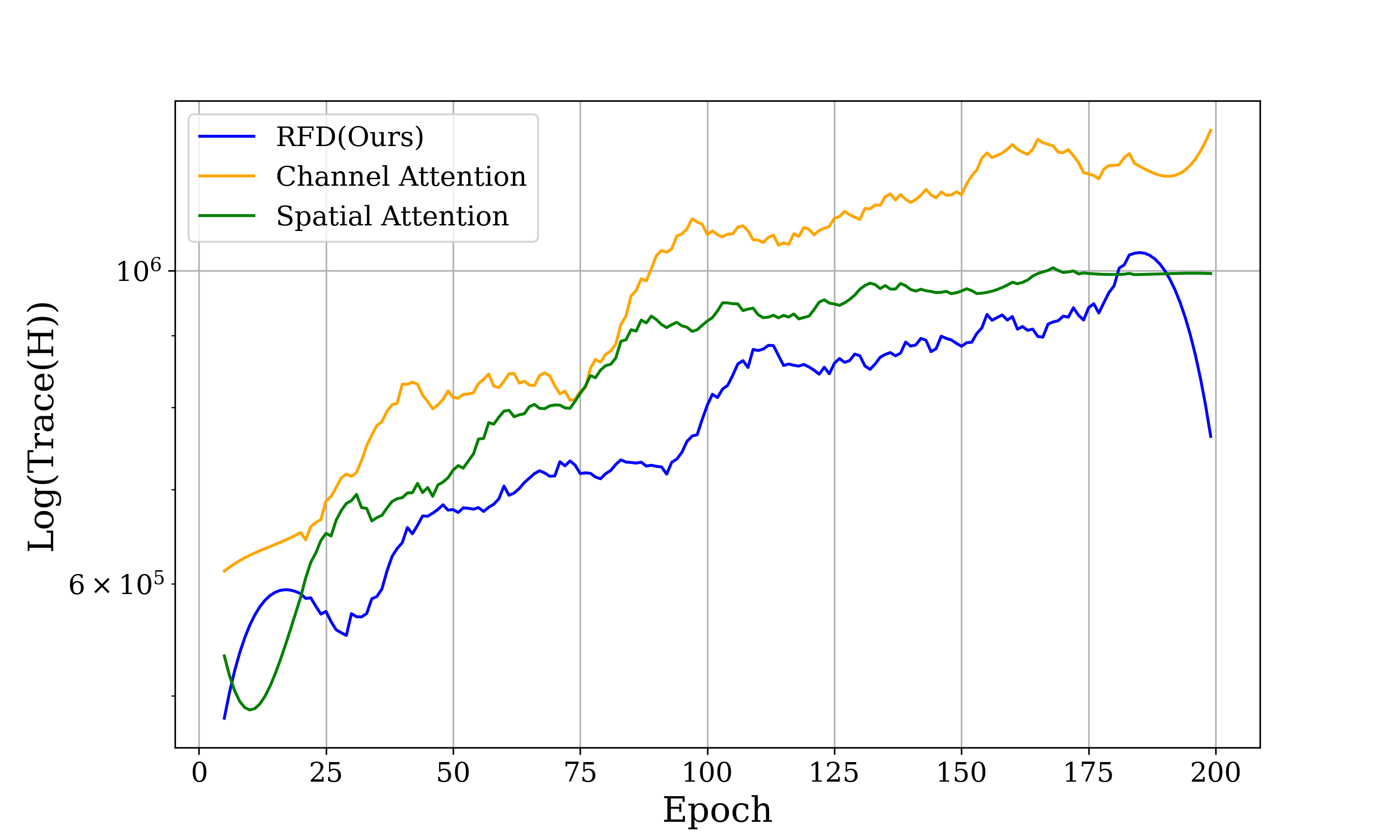}
        \caption{4-bit Quantization}
        \label{fig3b}
    \end{subfigure}
    
    \caption{Curvature Based on Hessian Trace in 3,4 bit Quantization.}
    \label{fig3}
\end{figure}

\subsection{Experimental Results}

The AKT method is a novel approach aimed at enhancing the effective training of quantized models. Therefore, AKT can be applied to all generative methods actively studied in the field of ZSQ. To validate the effectiveness of AKT, we applied the method to three popular generative approaches. First, AKT was applied to GDFQ\cite{22}, which originally proposed generative ZSQ. Second, we leveraged HAST\cite{16}, which improves the training difficulty of generated examples. Lastly, we used AdaDFQ\cite{18}, a generative model that enhances the generalization ability of quantized models. The experimental results are displayed in Table \ref{tab2}. Notably, AKT improved performance across all results except for the 5-bit GDFQ on the CIFAR-10 dataset. In particular, AdaDFQ+AKT achieved a remarkable 1.87\% performance improvement in 3-bit quantization. Additionally, our method shows increasingly pronounced performance improvements as the bit width decreases. This effect corresponds to a higher level of model compression. For instance, on CIFAR-10, the accuracy of AdaDFQ+AKT improved by 0.02\% at 5-bit, 0.33\% at 4-bit, and significantly by 1.87\% at 3-bit. This demonstrates that higher model compression reduces the information capacity of quantized models, and the AKT method effectively addresses this challenge.


In addition to validating the enhanced learning capabilities of our AKT method, we also conducted a comparative analysis against the latest state-of-the-art methods, as presented in Table \ref{tab3}. In this comparison, we applied our AKT method to the most recent approach with the available official code, AdaDFQ\cite{18}. The results demonstrate that training AdaDFQ with AKT leads to performance that matches or even surpasses that of TexQ and RIS, which previously outperformed AdaDFQ. Accordingly, to the best of our knowledge, our approach achieves state-of-the-art performance in the 3-bit and 5-bit scenarios, as evidenced by the CIFAR-10 results in Table \ref{tab3}. However, TexQ achieves significantly better performance than AdaDFQ on CIFAR-100. As a result, even when applying the AKT method to AdaDFQ, it is difficult to surpass TexQ's performance. Although we could not conduct experiments with AKT on TexQ due to the absence of its official code, we expect that applying AKT to TexQ would further enhance TexQ's performance.


\subsection{Ablation Study}

In this section, we conduct ablation studies to analyze the effectiveness of combining spatial and channel attention information for training quantized models. As shown in Table \ref{tab4}, integrating both types of attention proves to be much more effective for ZSQ compared to using either spatial or channel attention alone. In all 3, 4, and 5-bit environments, our proposed refined feature distillation outperforms both spatial and channel attention approaches. Notably, the experimental results clearly indicate that RFD performs better as the model bit-width decreases. In a 5-bit setting, RFD achieves a performance of 93.83\%, which is 0.21\% higher than the 93.62\% of channel attention. Furthermore, RFD outperforms channel attention by 0.28\% in a 4-bit setting. Remarkably, in a 3-bit environment, RFD shows a significant performance improvement of 0.67\%. These results demonstrate that our method is especially effective in low-bit models.

To further verify the effect of merging these two attention types in the refined feature distillation step, we perform curvature analysis based on the hessian trace\cite{40, 41, 42}. The hessian trace offers a reliable measure of training stability and generalization capability as it represents the second derivative of the loss function. Lower values in the hessian trace curvature imply improved performance and more stability. Figure \ref{fig3} provides a visualization under identical conditions for different losses. As shown in the Figure \ref{fig3}, our refined feature distillation exhibits the lowest hessian trace values in both 3-bit and 4-bit quantization settings, indicating the most stable learning process. In particular, under 3-bit quantization, both spatial and channel attention result in a large increase in curvature. In contrast, our method demonstrates significantly more stable learning compared to both approaches. This is because it more effectively distills the information contained in the feature maps. An additional insight from Figure \ref{fig3} is that our method is more effective for low-bit (3-bit) quantized models compared to high-bit (4-bit) quantized models. In Figure \ref{fig3b}, although our approach shows a more stable curvature compared to spatial and channel attention, the difference is not significant. However, in Figure \ref{fig3a}, our method demonstrates a remarkably stable learning process.



\section{Discussion}
\label{sec:discussion}

In conventional quantization, feature distillation approaches rely solely on either channel-attentive or spatial-attentive information from feature maps. This one-sided attention results in a significant loss of the original data in the feature maps. Our AKT method overcomes this limitation by merging channel and spatial attention maps before distillation, thereby effectively incorporating both types of features. This dual-perspective attention on feature maps is novel in ZSQ. Tables \ref{tab2} and \ref{tab4} highlight that our AKT method achieves remarkable performance. This implies that even with limited information capacity, low-bit quantized model has effectively acquired key knowledge during training. As a result of this evaluation, we verified that our method enhances the performance of the 3-bit quantized model. Furthermore, our approach reduces the performance degradation in 4- and 5-bit quantized models, which closely approximate the full-precision model.

However, our AKT method also has some limitations. Since the AKT method utilizes feature distillation, it is easily affected by the quality of input data. This is because the feature map used in feature distillation largely depends on not only the weight but also the input data. In ZSQ settings where training data is not available, generating data from a full-precision model is necessary. Consequently, the quality of the generated data can significantly impact the performance of the AKT method. Hence, we highlight the importance of exploring training methodologies for quantized models. Additionally, we emphasize the importance of investigating the combination of generative and learning methodologies.



\section{Conclusion}
\label{sec:conclusion}

The objective of this study is to reduce the information gap (i.e., quantization noise \cite{43}) between teacher models (full-precision models) and student models (quantized models) during knowledge distillation in ZSQ. To address this, we propose the AKT (Advanced Knowledge Transfer) method, which is designed to transfer key feature information from the teacher model to the quantized model. The AKT method comprises three key stages: 1) dual-information decomposition, 2) refined feature distillation, and 3) advanced knowledge transfer. Through this approach, quantized model can effectively acquire knowledge from the full-precision model, even with its limited information capacity. Experimental results demonstrate that our method outperforms a variety of existing ZSQ techniques. Notably, AKT is particularly effective in low-bit quantization settings, which often require further performance improvement. We also analyzed that feature information has a positive impact on quantization. Moreover, since AKT is a training-based approach, it can be widely applied across various generative methods. We expect that our work will stimulate further research in both the generative and learning aspects of ZSQ in the future. Furthermore, we hope that this research will contribute to the advancement of edge computing.


\section*{Acknowledgements}
This research was supported by the National Research Foundation (NRF) funded by the Korean government (MSIT) (No. RS-2023-00229822).
\nocite{langley00}


\begin{thebibliography}{64}
\providecommand{\natexlab}[1]{#1}
\providecommand{\url}[1]{\texttt{#1}}
\expandafter\ifx\csname urlstyle\endcsname\relax
  \providecommand{\doi}[1]{doi: #1}\else
  \providecommand{\doi}{doi: \begingroup \urlstyle{rm}\Url}\fi

\bibitem[Bai et~al.(2024)Bai, Yang, Chu, Wang, Liu, Chen, He, Mu, Cai, and
  Hu]{20}
Bai, J., Yang, Y., Chu, H., Wang, H., Liu, Z., Chen, R., He, X., Mu, L., Cai,
  C., and Hu, H.
\newblock Robustness-guided image synthesis for data-free quantization.
\newblock In \emph{Proceedings of the AAAI Conference on Artificial
  Intelligence}, volume~38, pp.\  10971--10979, 2024.

\bibitem[Becker et~al.(1988)Becker, Le~Cun, et~al.]{41}
Becker, S., Le~Cun, Y., et~al.
\newblock Improving the convergence of back-propagation learning with second
  order methods.
\newblock In \emph{Proceedings of the 1988 connectionist models summer school},
  pp.\  29--37, 1988.

\bibitem[Bottou et~al.(2018)Bottou, Curtis, and Nocedal]{42}
Bottou, L., Curtis, F.~E., and Nocedal, J.
\newblock Optimization methods for large-scale machine learning.
\newblock \emph{SIAM review}, 60\penalty0 (2):\penalty0 223--311, 2018.

\bibitem[Cai et~al.(2020)Cai, Yao, Dong, Gholami, Mahoney, and Keutzer]{9}
Cai, Y., Yao, Z., Dong, Z., Gholami, A., Mahoney, M.~W., and Keutzer, K.
\newblock Zeroq: A novel zero shot quantization framework.
\newblock In \emph{Proceedings of the IEEE/CVF conference on computer vision
  and pattern recognition}, pp.\  13169--13178, 2020.

\bibitem[Chen et~al.(2024{\natexlab{a}})Chen, Shao, Xu, Wang, Gao, Zhang, Qiao,
  and Luo]{56}
Chen, M., Shao, W., Xu, P., Wang, J., Gao, P., Zhang, K., Qiao, Y., and Luo, P.
\newblock Efficientqat: Efficient quantization-aware training for large
  language models.
\newblock \emph{arXiv preprint arXiv:2407.11062}, 2024{\natexlab{a}}.

\bibitem[Chen et~al.(2024{\natexlab{b}})Chen, Wang, Yan, Liu, Guan, and He]{19}
Chen, X., Wang, Y., Yan, R., Liu, Y., Guan, T., and He, Y.
\newblock Texq: zero-shot network quantization with texture feature
  distribution calibration.
\newblock \emph{Advances in Neural Information Processing Systems}, 36,
  2024{\natexlab{b}}.

\bibitem[Choi et~al.(2021)Choi, Hong, Park, Kim, and Lee]{14}
Choi, K., Hong, D., Park, N., Kim, Y., and Lee, J.
\newblock Qimera: Data-free quantization with synthetic boundary supporting
  samples.
\newblock \emph{Advances in Neural Information Processing Systems},
  34:\penalty0 14835--14847, 2021.

\bibitem[Choi et~al.(2022)Choi, Lee, Hong, Yu, Park, Kim, and Lee]{12}
Choi, K., Lee, H.~Y., Hong, D., Yu, J., Park, N., Kim, Y., and Lee, J.
\newblock It's all in the teacher: Zero-shot quantization brought closer to the
  teacher.
\newblock In \emph{Proceedings of the IEEE/CVF Conference on Computer Vision
  and Pattern Recognition}, pp.\  8311--8321, 2022.

\bibitem[Choi et~al.(2016)Choi, El-Khamy, and Lee]{26}
Choi, Y., El-Khamy, M., and Lee, J.
\newblock Towards the limit of network quantization.
\newblock \emph{arXiv preprint arXiv:1612.01543}, 2016.

\bibitem[Choi et~al.(2020)Choi, Choi, El-Khamy, and Lee]{63}
Choi, Y., Choi, J., El-Khamy, M., and Lee, J.
\newblock Data-free network quantization with adversarial knowledge
  distillation.
\newblock In \emph{Proceedings of the IEEE/CVF Conference on Computer Vision
  and Pattern Recognition Workshops}, pp.\  710--711, 2020.

\bibitem[Courbariaux et~al.(2015)Courbariaux, Bengio, and David]{6}
Courbariaux, M., Bengio, Y., and David, J.-P.
\newblock Binaryconnect: Training deep neural networks with binary weights
  during propagations.
\newblock \emph{Advances in neural information processing systems}, 28, 2015.

\bibitem[Ding et~al.(2024)Ding, Feng, Chen, Guo, and Liu]{51}
Ding, Y., Feng, W., Chen, C., Guo, J., and Liu, X.
\newblock Reg-ptq: Regression-specialized post-training quantization for fully
  quantized object detector.
\newblock In \emph{Proceedings of the IEEE/CVF Conference on Computer Vision
  and Pattern Recognition}, pp.\  16174--16184, 2024.

\bibitem[Fan et~al.(2020)Fan, Stock, Graham, Grave, Gribonval, Jegou, and
  Joulin]{7}
Fan, A., Stock, P., Graham, B., Grave, E., Gribonval, R., Jegou, H., and
  Joulin, A.
\newblock Training with quantization noise for extreme model compression.
\newblock \emph{arXiv preprint arXiv:2004.07320}, 2020.

\bibitem[Francy \& Singh(2024)Francy and Singh]{47}
Francy, S. and Singh, R.
\newblock Edge ai: Evaluation of model compression techniques for convolutional
  neural networks.
\newblock \emph{arXiv preprint arXiv:2409.02134}, 2024.

\bibitem[Gholami et~al.(2022)Gholami, Kim, Dong, Yao, Mahoney, and Keutzer]{27}
Gholami, A., Kim, S., Dong, Z., Yao, Z., Mahoney, M.~W., and Keutzer, K.
\newblock A survey of quantization methods for efficient neural network
  inference.
\newblock In \emph{Low-Power Computer Vision}, pp.\  291--326. Chapman and
  Hall/CRC, 2022.

\bibitem[Gu \& Dao(2023)Gu and Dao]{32}
Gu, A. and Dao, T.
\newblock Mamba: Linear-time sequence modeling with selective state spaces.
\newblock \emph{arXiv preprint arXiv:2312.00752}, 2023.

\bibitem[Guo et~al.(2022)Guo, Qiu, Leng, Gao, Zhang, Liu, Yang, Zhu, and
  Guo]{11}
Guo, C., Qiu, Y., Leng, J., Gao, X., Zhang, C., Liu, Y., Yang, F., Zhu, Y., and
  Guo, M.
\newblock Squant: On-the-fly data-free quantization via diagonal hessian
  approximation.
\newblock \emph{arXiv preprint arXiv:2202.07471}, 2022.

\bibitem[Han et~al.(2015)Han, Mao, and Dally]{2}
Han, S., Mao, H., and Dally, W.~J.
\newblock Deep compression: Compressing deep neural networks with pruning,
  trained quantization and huffman coding.
\newblock \emph{arXiv preprint arXiv:1510.00149}, 2015.

\bibitem[Han et~al.(2016)Han, Liu, Mao, Pu, Pedram, Horowitz, and Dally]{23}
Han, S., Liu, X., Mao, H., Pu, J., Pedram, A., Horowitz, M.~A., and Dally,
  W.~J.
\newblock Eie: Efficient inference engine on compressed deep neural network.
\newblock \emph{ACM SIGARCH Computer Architecture News}, 44\penalty0
  (3):\penalty0 243--254, 2016.

\bibitem[He et~al.(2016)He, Zhang, Ren, and Sun]{38}
He, K., Zhang, X., Ren, S., and Sun, J.
\newblock Deep residual learning for image recognition.
\newblock In \emph{Proceedings of the IEEE conference on computer vision and
  pattern recognition}, pp.\  770--778, 2016.

\bibitem[He et~al.(2023)He, Liu, Wu, Zhou, and Zhuang]{58}
He, Y., Liu, J., Wu, W., Zhou, H., and Zhuang, B.
\newblock Efficientdm: Efficient quantization-aware fine-tuning of low-bit
  diffusion models.
\newblock \emph{arXiv preprint arXiv:2310.03270}, 2023.

\bibitem[Heo et~al.(2019)Heo, Kim, Yun, Park, Kwak, and Choi]{36}
Heo, B., Kim, J., Yun, S., Park, H., Kwak, N., and Choi, J.~Y.
\newblock A comprehensive overhaul of feature distillation.
\newblock In \emph{Proceedings of the IEEE/CVF international conference on
  computer vision}, pp.\  1921--1930, 2019.

\bibitem[Hinton(2015)]{1}
Hinton, G.
\newblock Distilling the knowledge in a neural network.
\newblock \emph{arXiv preprint arXiv:1503.02531}, 2015.

\bibitem[Hong \& Choi(2023)Hong and Choi]{31}
Hong, I. and Choi, C.
\newblock Knowledge distillation vulnerability of deit through cnn adversarial
  attack.
\newblock \emph{Neural Computing and Applications}, pp.\  1--11, 2023.

\bibitem[Hong \& Lee(2024)Hong and Lee]{64}
Hong, I. and Lee, S.
\newblock Exploring synergy of denoising and distillation: Novel method for
  efficient adversarial defense.
\newblock \emph{Applied Sciences}, 14\penalty0 (23):\penalty0 10872, 2024.

\bibitem[Hong et~al.(2023)Hong, Choi, Kim, and Choi]{30}
Hong, I.-p., Choi, G.-h., Kim, P.-k., and Choi, C.
\newblock Security verification software platform of data-efficient image
  transformer based on fast gradient sign method.
\newblock In \emph{Proceedings of the 38th ACM/SIGAPP Symposium on Applied
  Computing}, pp.\  1669--1672, 2023.

\bibitem[Howard(2017)]{48}
Howard, A.~G.
\newblock Mobilenets: Efficient convolutional neural networks for mobile vision
  applications.
\newblock \emph{arXiv preprint arXiv:1704.04861}, 2017.

\bibitem[Hua et~al.(2023)Hua, Li, Wang, Dong, Li, and Cao]{45}
Hua, H., Li, Y., Wang, T., Dong, N., Li, W., and Cao, J.
\newblock Edge computing with artificial intelligence: A machine learning
  perspective.
\newblock \emph{ACM Computing Surveys}, 55\penalty0 (9):\penalty0 1--35, 2023.

\bibitem[Hubara et~al.(2018)Hubara, Courbariaux, Soudry, El-Yaniv, and
  Bengio]{3}
Hubara, I., Courbariaux, M., Soudry, D., El-Yaniv, R., and Bengio, Y.
\newblock Quantized neural networks: Training neural networks with low
  precision weights and activations.
\newblock \emph{Journal of Machine Learning Research}, 18\penalty0
  (187):\penalty0 1--30, 2018.

\bibitem[Krishnamoorthi(2018)]{25}
Krishnamoorthi, R.
\newblock Quantizing deep convolutional networks for efficient inference: A
  whitepaper.
\newblock \emph{arXiv preprint arXiv:1806.08342}, 2018.

\bibitem[Krizhevsky et~al.(2009)Krizhevsky, Hinton, et~al.]{37}
Krizhevsky, A., Hinton, G., et~al.
\newblock Learning multiple layers of features from tiny images.
\newblock 2009.

\bibitem[Kulkarni et~al.(2021)Kulkarni, Meena, Gurlahosur, Benagi, Kashyap,
  Ansari, and Karnam]{46}
Kulkarni, U., Meena, S., Gurlahosur, S.~V., Benagi, P., Kashyap, A., Ansari,
  A., and Karnam, V.
\newblock Ai model compression for edge devices using optimization techniques.
\newblock In \emph{Modern Approaches in Machine Learning and Cognitive Science:
  A Walkthrough: Latest Trends in AI, Volume 2}, pp.\  227--240. Springer,
  2021.

\bibitem[Kuzmin et~al.(2024)Kuzmin, Nagel, Van~Baalen, Behboodi, and
  Blankevoort]{60}
Kuzmin, A., Nagel, M., Van~Baalen, M., Behboodi, A., and Blankevoort, T.
\newblock Pruning vs quantization: which is better?
\newblock \emph{Advances in neural information processing systems}, 36, 2024.

\bibitem[Li et~al.(2023)Li, Wu, Lv, Liao, Li, Zhang, Han, and Tan]{16}
Li, H., Wu, X., Lv, F., Liao, D., Li, T.~H., Zhang, Y., Han, B., and Tan, M.
\newblock Hard sample matters a lot in zero-shot quantization.
\newblock In \emph{Proceedings of the IEEE/CVF conference on Computer Vision
  and Pattern Recognition}, pp.\  24417--24426, 2023.

\bibitem[Liang et~al.(2021)Liang, Glossner, Wang, Shi, and Zhang]{61}
Liang, T., Glossner, J., Wang, L., Shi, S., and Zhang, X.
\newblock Pruning and quantization for deep neural network acceleration: A
  survey.
\newblock \emph{Neurocomputing}, 461:\penalty0 370--403, 2021.

\bibitem[Lin et~al.(2023)Lin, Peng, Li, Tan, Ren, Xiao, and Pu]{43}
Lin, C., Peng, B., Li, Z., Tan, W., Ren, Y., Xiao, J., and Pu, S.
\newblock Bit-shrinking: Limiting instantaneous sharpness for improving
  post-training quantization.
\newblock In \emph{Proceedings of the IEEE/CVF Conference on Computer Vision
  and Pattern Recognition}, pp.\  16196--16205, 2023.

\bibitem[Lin et~al.(2016)Lin, Talathi, and Annapureddy]{4}
Lin, D., Talathi, S., and Annapureddy, S.
\newblock Fixed point quantization of deep convolutional networks.
\newblock In \emph{International conference on machine learning}, pp.\
  2849--2858. PMLR, 2016.

\bibitem[Liu et~al.(2023)Liu, Sun, and Katto]{50}
Liu, J., Sun, H., and Katto, J.
\newblock Learned image compression with mixed transformer-cnn architectures.
\newblock In \emph{Proceedings of the IEEE/CVF conference on computer vision
  and pattern recognition}, pp.\  14388--14397, 2023.

\bibitem[Liu \& Chen(2024)Liu and Chen]{52}
Liu, K. and Chen, N.
\newblock Ptq-so: A scale optimization-based approach for post-training
  quantization of edge computing.
\newblock In \emph{2024 27th International Conference on Computer Supported
  Cooperative Work in Design (CSCWD)}, pp.\  2078--2083. IEEE, 2024.

\bibitem[McEnroe et~al.(2022)McEnroe, Wang, and Liyanage]{44}
McEnroe, P., Wang, S., and Liyanage, M.
\newblock A survey on the convergence of edge computing and ai for uavs:
  Opportunities and challenges.
\newblock \emph{IEEE Internet of Things Journal}, 9\penalty0 (17):\penalty0
  15435--15459, 2022.

\bibitem[Mentzer et~al.(2020)Mentzer, Toderici, Tschannen, and Agustsson]{24}
Mentzer, F., Toderici, G.~D., Tschannen, M., and Agustsson, E.
\newblock High-fidelity generative image compression.
\newblock \emph{Advances in Neural Information Processing Systems},
  33:\penalty0 11913--11924, 2020.

\bibitem[Nagel et~al.(2019)Nagel, Baalen, Blankevoort, and Welling]{62}
Nagel, M., Baalen, M.~v., Blankevoort, T., and Welling, M.
\newblock Data-free quantization through weight equalization and bias
  correction.
\newblock In \emph{Proceedings of the IEEE/CVF International Conference on
  Computer Vision}, pp.\  1325--1334, 2019.

\bibitem[Park et~al.(2023)Park, Hong, Poudel, and Choi]{35}
Park, H.-C., Hong, I.-P., Poudel, S., and Choi, C.
\newblock Data augmentation based on generative adversarial networks for
  endoscopic image classification.
\newblock \emph{IEEE Access}, 11:\penalty0 49216--49225, 2023.

\bibitem[Paszke et~al.(2019)Paszke, Gross, Massa, Lerer, Bradbury, Chanan,
  Killeen, Lin, Gimelshein, Antiga, et~al.]{39}
Paszke, A., Gross, S., Massa, F., Lerer, A., Bradbury, J., Chanan, G., Killeen,
  T., Lin, Z., Gimelshein, N., Antiga, L., et~al.
\newblock Pytorch: An imperative style, high-performance deep learning library.
\newblock \emph{Advances in neural information processing systems}, 32, 2019.

\bibitem[Pearlmutter(1994)]{40}
Pearlmutter, B.~A.
\newblock Fast exact multiplication by the hessian.
\newblock \emph{Neural computation}, 6\penalty0 (1):\penalty0 147--160, 1994.

\bibitem[Qian et~al.(2023{\natexlab{a}})Qian, Wang, Hong, and Wang]{17}
Qian, B., Wang, Y., Hong, R., and Wang, M.
\newblock Rethinking data-free quantization as a zero-sum game.
\newblock In \emph{Proceedings of the AAAI conference on artificial
  intelligence}, volume~37, pp.\  9489--9497, 2023{\natexlab{a}}.

\bibitem[Qian et~al.(2023{\natexlab{b}})Qian, Wang, Hong, and Wang]{18}
Qian, B., Wang, Y., Hong, R., and Wang, M.
\newblock Adaptive data-free quantization.
\newblock In \emph{Proceedings of the IEEE/CVF Conference on Computer Vision
  and Pattern Recognition}, pp.\  7960--7968, 2023{\natexlab{b}}.

\bibitem[Sandler et~al.(2018)Sandler, Howard, Zhu, Zhmoginov, and Chen]{49}
Sandler, M., Howard, A., Zhu, M., Zhmoginov, A., and Chen, L.-C.
\newblock Mobilenetv2: Inverted residuals and linear bottlenecks.
\newblock In \emph{Proceedings of the IEEE conference on computer vision and
  pattern recognition}, pp.\  4510--4520, 2018.

\bibitem[Sevsay \& Akag{\"u}nd{\"u}z(2024)Sevsay and Akag{\"u}nd{\"u}z]{54}
Sevsay, B. and Akag{\"u}nd{\"u}z, E.
\newblock Infrared domain adaptation with zero-shot quantization.
\newblock \emph{arXiv preprint arXiv:2408.13925}, 2024.

\bibitem[Shang et~al.(2023)Shang, Yuan, Xie, Wu, and Yan]{8}
Shang, Y., Yuan, Z., Xie, B., Wu, B., and Yan, Y.
\newblock Post-training quantization on diffusion models.
\newblock In \emph{Proceedings of the IEEE/CVF conference on computer vision
  and pattern recognition}, pp.\  1972--1981, 2023.

\bibitem[Shi et~al.(2023)Shi, Lu, and Ma]{53}
Shi, J., Lu, M., and Ma, Z.
\newblock Rate-distortion optimized post-training quantization for learned
  image compression.
\newblock \emph{IEEE Transactions on Circuits and Systems for Video
  Technology}, 2023.

\bibitem[Stanton et~al.(2021)Stanton, Izmailov, Kirichenko, Alemi, and
  Wilson]{28}
Stanton, S., Izmailov, P., Kirichenko, P., Alemi, A.~A., and Wilson, A.~G.
\newblock Does knowledge distillation really work?
\newblock \emph{Advances in Neural Information Processing Systems},
  34:\penalty0 6906--6919, 2021.

\bibitem[Wang et~al.(2024{\natexlab{a}})Wang, Chen, Liu, Chen, Lin, Han, and
  Ding]{33}
Wang, A., Chen, H., Liu, L., Chen, K., Lin, Z., Han, J., and Ding, G.
\newblock Yolov10: Real-time end-to-end object detection.
\newblock \emph{arXiv preprint arXiv:2405.14458}, 2024{\natexlab{a}}.

\bibitem[Wang et~al.(2024{\natexlab{b}})Wang, Zeng, Liang, Xing, Xu, and
  Xu]{34}
Wang, Y., Zeng, Y., Liang, J., Xing, X., Xu, J., and Xu, X.
\newblock Retrievalmmt: Retrieval-constrained multi-modal prompt learning for
  multi-modal machine translation.
\newblock In \emph{Proceedings of the 2024 International Conference on
  Multimedia Retrieval}, pp.\  860--868, 2024{\natexlab{b}}.

\bibitem[Wu et~al.(2020)Wu, Judd, Zhang, Isaev, and Micikevicius]{21}
Wu, H., Judd, P., Zhang, X., Isaev, M., and Micikevicius, P.
\newblock Integer quantization for deep learning inference: Principles and
  empirical evaluation.
\newblock \emph{arXiv preprint arXiv:2004.09602}, 2020.

\bibitem[Xu et~al.(2020{\natexlab{a}})Xu, Li, Zhuang, Liu, Cao, Liang, and
  Tan]{10}
Xu, S., Li, H., Zhuang, B., Liu, J., Cao, J., Liang, C., and Tan, M.
\newblock Generative low-bitwidth data free quantization.
\newblock In \emph{Computer Vision--ECCV 2020: 16th European Conference,
  Glasgow, UK, August 23--28, 2020, Proceedings, Part XII 16}, pp.\  1--17.
  Springer, 2020{\natexlab{a}}.

\bibitem[Xu et~al.(2020{\natexlab{b}})Xu, Li, Zhuang, Liu, Cao, Liang, and
  Tan]{22}
Xu, S., Li, H., Zhuang, B., Liu, J., Cao, J., Liang, C., and Tan, M.
\newblock Generative low-bitwidth data free quantization.
\newblock In \emph{Computer Vision--ECCV 2020: 16th European Conference,
  Glasgow, UK, August 23--28, 2020, Proceedings, Part XII 16}, pp.\  1--17.
  Springer, 2020{\natexlab{b}}.

\bibitem[Yuan et~al.(2020)Yuan, Tay, Li, Wang, and Feng]{29}
Yuan, L., Tay, F.~E., Li, G., Wang, T., and Feng, J.
\newblock Revisiting knowledge distillation via label smoothing regularization.
\newblock In \emph{Proceedings of the IEEE/CVF conference on computer vision
  and pattern recognition}, pp.\  3903--3911, 2020.

\bibitem[Zhang \& Chung(2024)Zhang and Chung]{55}
Zhang, R. and Chung, A.~C.
\newblock Efficientq: An efficient and accurate post-training neural network
  quantization method for medical image segmentation.
\newblock \emph{Medical Image Analysis}, 97:\penalty0 103277, 2024.

\bibitem[Zhang et~al.(2021)Zhang, Qin, Ding, Gong, Yan, Tao, Li, Yu, and
  Liu]{13}
Zhang, X., Qin, H., Ding, Y., Gong, R., Yan, Q., Tao, R., Li, Y., Yu, F., and
  Liu, X.
\newblock Diversifying sample generation for accurate data-free quantization.
\newblock In \emph{Proceedings of the IEEE/CVF conference on computer vision
  and pattern recognition}, pp.\  15658--15667, 2021.

\bibitem[Zhao \& Zhao(2024)Zhao and Zhao]{57}
Zhao, K. and Zhao, M.
\newblock Self-supervised quantization-aware knowledge distillation.
\newblock \emph{arXiv preprint arXiv:2403.11106}, 2024.

\bibitem[Zhong et~al.(2022)Zhong, Lin, Nan, Liu, Zhang, Tian, and Ji]{15}
Zhong, Y., Lin, M., Nan, G., Liu, J., Zhang, B., Tian, Y., and Ji, R.
\newblock Intraq: Learning synthetic images with intra-class heterogeneity for
  zero-shot network quantization.
\newblock In \emph{Proceedings of the IEEE/CVF Conference on Computer Vision
  and Pattern Recognition}, pp.\  12339--12348, 2022.

\bibitem[Zhou et~al.(2016)Zhou, Wu, Ni, Zhou, Wen, and Zou]{5}
Zhou, S., Wu, Y., Ni, Z., Zhou, X., Wen, H., and Zou, Y.
\newblock Dorefa-net: Training low bitwidth convolutional neural networks with
  low bitwidth gradients.
\newblock \emph{arXiv preprint arXiv:1606.06160}, 2016.

\bibitem[Zoph(2016)]{59}
Zoph, B.
\newblock Neural architecture search with reinforcement learning.
\newblock \emph{arXiv preprint arXiv:1611.01578}, 2016.

\end{thebibliography}

\end{document}